\documentclass[runningheads]{llncs}
\usepackage[T1]{fontenc}
\usepackage{booktabs}
\usepackage{graphicx}
\usepackage{subcaption}
\usepackage[export]{adjustbox} 
\usepackage{hyperref}
\usepackage{graphicx}
\begin{document}
\title{Exploring Self-Supervised Learning with U-Net Masked Autoencoders and EfficientNet-B7 for Improved Gastrointestinal Abnormality Classification in Video Capsule Endoscopy}
%
%
\author{Vamshi Krishna Kancharla\inst{1} \and
Pavan Kumar Kaveti\inst{2} \and
Dasari Naga Raju\inst{1}}
\authorrunning{F. Kancharla VK et al.}
%
\institute{International Institute of Information Technology, Bangalore  \and
Indian Institute of Technology, Hyderabad\\
\email{\{kancharla.vamshi\}@iiitb.ac.in}}
\maketitle              
\begin{abstract}
Video Capsule Endoscopy (VCE) has become an indispensable diagnostic tool for gastrointestinal (GI) disorders due to its non-invasive nature and ability to capture high-resolution images of the small intestine. However, the enormous volume of data generated during a single procedure makes manual inspection labor-intensive, time-consuming, and prone to inter-observer variability. Automated analysis using deep learning offers a promising solution, but its effectiveness is often limited by data imbalance and the high cost of labeled medical data. In this work, we propose a novel framework that combines self-supervised learning through a U-Net-based masked autoencoder with supervised feature extraction using EfficientNet-B7 for multi-class abnormality classification in VCE images. The U-Net model is first trained in a self-supervised manner using Gaussian noise removal and masked reconstruction to learn robust visual representations without requiring annotations. The learned encoder features are then fused with EfficientNet-B7 features to form a rich, discriminative representation for classification. We evaluate our approach on the Capsule Vision 2024 Challenge dataset consisting of ten abnormality classes and a dominant normal class. Experimental results demonstrate that the proposed fusion framework achieves a validation accuracy of 94\%, outperforming standalone architectures and attention-based fusion variants. The study highlights the effectiveness of self-supervised representation learning and feature fusion in addressing class imbalance and improving diagnostic accuracy in real-world medical imaging scenarios.
\textbf{Code:} \href{https://github.com/kancharlavamshi/Self-Supervised-U-Net-Masked-Autoencoders-and-EfficientNet-B7-for-Gastrointestinal-Abnormality-Class}{GitHub}

\keywords{Video Capsule Endoscopy \and Autoencoders\and Self-Supervised Learning\and Masked Image Modeling} \and UNet
\end{abstract}
\section{Introduction}
Gastrointestinal (GI) diseases represent a significant global health burden, affecting millions of individuals each year and contributing substantially to morbidity and healthcare costs. Conditions such as Crohn’s disease, celiac disease, gastrointestinal bleeding, ulcers, polyps, and intestinal cancers often require early detection for effective treatment and improved patient outcomes. Traditional endoscopic procedures, although effective, are invasive and uncomfortable for patients. Video Capsule Endoscopy (VCE) has emerged as a revolutionary, patient-friendly alternative that allows non-invasive visualization of the entire small intestine. A swallowable capsule equipped with a miniature camera captures tens of thousands of images as it traverses the GI tract, providing clinicians with valuable diagnostic information.

Despite its clinical advantages, VCE introduces new challenges in data management and interpretation. A single examination can generate between 57,000 and over 1,000,000 frames, depending on the duration of the recording and frame rate. Manual review of these frames by gastroenterologists is extremely time-consuming, often requiring two to three hours per patient. This process is not only labor-intensive but also subject to fatigue, human bias, and inter-observer variability, which can lead to missed lesions or false positives. Artifacts such as bubbles, food debris, intestinal folds, and motion blur further complicate accurate interpretation. Moreover, the growing demand for GI diagnostics has created a shortage of trained specialists, resulting in delays in diagnosis and treatment.

To address these limitations, there has been a rapid increase in research on automated VCE analysis using artificial intelligence (AI) and deep learning techniques. Convolutional neural networks (CNNs) have demonstrated impressive performance in various medical imaging tasks, including lesion detection, polyp segmentation, and disease classification. However, most existing approaches rely heavily on supervised learning, which requires large amounts of annotated data. In the medical domain, labeling is expensive, time-consuming, and requires expert knowledge, making it difficult to scale. Furthermore, VCE datasets are typically highly imbalanced, with normal frames vastly outnumbering abnormal ones. This imbalance often leads to biased models that perform well on the majority class but poorly on rare yet clinically important abnormalities.

Self-supervised learning has recently gained attention as a powerful paradigm for learning meaningful representations from unlabeled data. By designing pretext tasks such as image reconstruction, masking, or denoising, models can learn rich visual features that transfer well to downstream tasks. Masked autoencoders, in particular, have shown strong performance in learning scalable and generalizable representations. U-Net architectures, originally developed for biomedical segmentation, are especially well-suited for reconstruction tasks due to their encoder–decoder structure and skip connections, which preserve spatial details.

In parallel, EfficientNet has emerged as one of the most effective CNN architectures for image classification, achieving state-of-the-art performance with fewer parameters through compound scaling of depth, width, and resolution. EfficientNet-B7, the largest variant, has demonstrated excellent generalization across diverse visual domains.

In this paper, we propose a novel framework that integrates the strengths of self-supervised learning and supervised classification by combining a U-Net-based masked autoencoder with EfficientNet-B7 through feature fusion. The U-Net is first trained in a self-supervised manner using masked inputs and Gaussian noise removal to learn robust, high-level representations. These learned features are then fused with EfficientNet-B7 features and passed through fully connected layers for multi-class classification. Our goal is to develop a vendor-independent, generalized AI model capable of automatically classifying ten clinically relevant VCE abnormalities: angioectasia, bleeding, erosion, erythema, foreign body, lymphangiectasia, polyp, ulcer, worms, and normal. By leveraging self-supervised learning and feature fusion, we aim to improve diagnostic accuracy, reduce dependence on labeled data, and address the challenges posed by class imbalance in real-world VCE datasets.

\section{Related Work}

Early automation in VCErelied on manual feature extraction, such as color histograms and texture descriptors, paired with standard classifiers \cite{okeke2025}. These methods, however, lack the generalization power needed for diverse patient data. The CNNs have since replaced hand-crafted features, becoming the standard for medical image analysis \cite{soffer2025}. Among these, EfficientNet-B7 stands out for its balance of accuracy and computational efficiency, achieving state-of-the-art results in fine-grained classification tasks \cite{togacar2022}. Several recent studies confirm that fusing EfficientNet features with other architectures improves detection rates for bleeding and ulcers \cite{kundu2022} \cite{khan2022}.

A major bottleneck in these supervised approaches is the dependency on large, annotated datasets, which are costly and difficult to obtain in the medical domain \cite{dataset2024}. To address this, researchers are increasingly turning to Self-Supervised Learning (SSL). By solving pretext tasks—such as image reconstruction models can learn robust feature representations without explicit labels.

It has been demonstrated that scalable visual features can be learned by masking a high percentage of the image and optimizing for the reconstruction of occluded regions \cite{he2022}. While this Masked Image Modeling technique was popularized by Vision Transformers\cite{caron2021}, it is equally effective for CNN-based architectures in medical imaging \cite{endovit2024}. The U-Net, originally designed for segmentation, is particularly well-suited for this reconstruction task due to its skip connections, which preserve spatial details essential for identifying small lesions.

Current literature typically treats self-supervised reconstruction and supervised classification as separate tasks. Fusion methods simply concatenate features from two supervised networks \cite{togacar2022}. The proposed work diverges from this trend by integrating the reconstruction capabilities of a self-supervised U-Net with the discriminative power of EfficientNet-B7. This hybrid approach leverages the vast amount of unlabeled VCE data to improve classification performance on the unbalanced Capsule Vision 2024 challenge dataset \cite{capsule2024}.

\section{Dataset and Preprocessing}

A critical aspect of developing robust and clinically relevant models for VCEanalysis is the use of well-curated and representative datasets. In this work, all experiments are conducted on the Capsule Vision 2024 Challenge dataset, which is one of the most comprehensive publicly available datasets for multi-class abnormality classification in VCE images. The dataset was released as part of the international Capsule Vision Challenge with the goal of fostering the development of generalized, vendor-independent AI models for automated gastrointestinal diagnosis.

\begin{figure}[htbp]
    \centering
    \newcommand{\gridwidth}{0.19\linewidth}

    \begin{subfigure}[b]{\gridwidth}
        \centering
        \textbf{\scriptsize Angioectasia} \\[3pt]  
        \includegraphics[width=\linewidth, keepaspectratio]{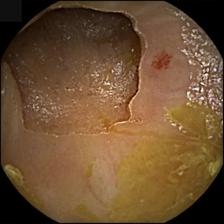}
    \end{subfigure}
    \hfill
    \begin{subfigure}[b]{\gridwidth}
        \centering
        \textbf{\scriptsize Bleeding} \\[3pt]
        \includegraphics[width=\linewidth, keepaspectratio]{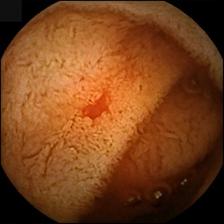}
    \end{subfigure}
    \hfill
    \begin{subfigure}[b]{\gridwidth}
        \centering
        \textbf{\scriptsize Erosion} \\[3pt]
        \includegraphics[width=\linewidth, keepaspectratio]{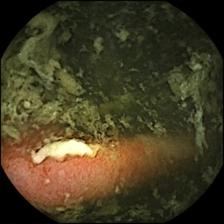}
    \end{subfigure}
    \hfill
    \begin{subfigure}[b]{\gridwidth}
        \centering
        \textbf{\scriptsize Erythema} \\[3pt]
        \includegraphics[width=\linewidth, keepaspectratio]{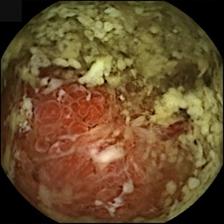}
    \end{subfigure}
    \hfill
    \begin{subfigure}[b]{\gridwidth}
        \centering
        \textbf{\scriptsize Foreign Body} \\[3pt]
        \includegraphics[width=\linewidth, keepaspectratio]{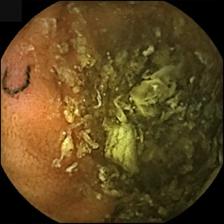}
    \end{subfigure}

    \vspace{1em} 

    \begin{subfigure}[b]{\gridwidth}
        \centering
        \resizebox{\linewidth}{!}{\textbf{\scriptsize Lymphangiectasia}} \\[3pt]
        \includegraphics[width=\linewidth, keepaspectratio]{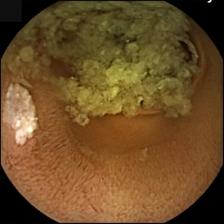}
    \end{subfigure}
    \hfill
    \begin{subfigure}[b]{\gridwidth}
        \centering
        \textbf{\scriptsize Normal} \\[3pt]
        \includegraphics[width=\linewidth, keepaspectratio]{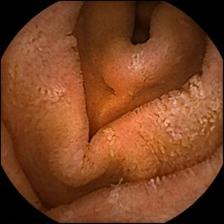}
    \end{subfigure}
    \hfill
    \begin{subfigure}[b]{\gridwidth}
        \centering
        \textbf{\scriptsize Polyp} \\[3pt]
        \includegraphics[width=\linewidth, keepaspectratio]{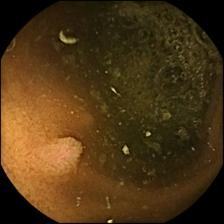}
    \end{subfigure}
    \hfill
    \begin{subfigure}[b]{\gridwidth}
        \centering
        \textbf{\scriptsize Ulcer} \\[3pt]
        \includegraphics[width=\linewidth, keepaspectratio]{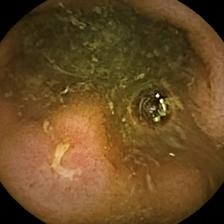}
    \end{subfigure}
    \hfill
    \begin{subfigure}[b]{\gridwidth}
        \centering
        \textbf{\scriptsize Worms} \\[3pt]
        \includegraphics[width=\linewidth, keepaspectratio]{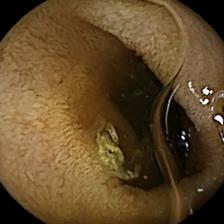}
    \end{subfigure}

    \caption{Representative example frames from the dataset. }
    \label{fig:dataset_samples}
\end{figure}

The dataset consists of high-resolution VCE frames collected from multiple patients and diverse clinical settings, ensuring variability in imaging conditions, patient anatomy, illumination, and artifact presence. Each image is annotated by expert gastroenterologists into one of ten clinically relevant categories: angioectasia, bleeding, erosion, erythema, foreign body, lymphangiectasia, polyp, ulcer, worms, and normal shown in Figure~\ref{fig:dataset_samples} . These categories cover a broad spectrum of gastrointestinal pathologies and represent common diagnostic targets in routine clinical practice.
One of the defining characteristics of the dataset is its severe class imbalance. The ``Normal'' class constitutes the majority of samples, while several abnormal classes such as worms, erythema, and angioectasia are comparatively rare. This imbalance reflects real-world clinical distributions, where pathological findings occur much less frequently than normal tissue. However, it also poses a significant challenge for supervised learning models, which tend to be biased toward majority classes and may underperform on rare but clinically critical abnormalities.

The dataset is split into training and validation subsets as defined by the challenge organizers. We strictly adhere to the official splits and competition rules, ensuring that no external data or annotations are used. This is essential for maintaining fair comparison and reproducibility. The total number of validation samples is 16,132, with class-wise distribution as reported in Table~\ref{tab:class_metrics} of the original paper, where the normal class alone accounts for over 12,000 images.

Before feeding the images into the network, several preprocessing steps are applied. First, all images are resized to match the input resolution required by EfficientNet-B7. Color normalization is performed to reduce inter-patient and inter-device variability. Since VCE images often suffer from illumination changes, specular highlights, and color distortions due to intestinal fluids, normalization helps stabilize training and improves convergence.

To improve generalization and mitigate overfitting, we apply standard data augmentation techniques during training. These include random horizontal and vertical flips, small rotations, zooming, and brightness adjustments. Augmentation is particularly important given the class imbalance, as it increases the effective diversity of minority class samples and encourages the model to learn invariant features.

For the self-supervised pretraining stage, both labeled and unlabeled images from the training set are used without their annotations. This allows the U-Net masked autoencoder to exploit the full data distribution and learn structural patterns of the gastrointestinal tract independent of class labels. Gaussian noise is injected into the images during this phase to create corrupted inputs for reconstruction. The target outputs are the original clean images, enabling the model to learn denoising and inpainting capabilities.

The supervised classification stage uses the labeled training subset. The reconstructed images from the U-Net branch and the original images are fed into EfficientNet-B7 and the U-Net encoder respectively, and their features are fused for final prediction. The validation set is used exclusively for evaluation and model selection.

By using the Capsule Vision 2024 dataset and adhering strictly to its protocols, we ensure that our results are directly comparable with other state-of-the-art methods and that the proposed approach is evaluated under realistic clinical conditions.

\section{Proposed Methodology}
\begin{figure}[htbp]
    \centering
    \includegraphics[width=\linewidth]{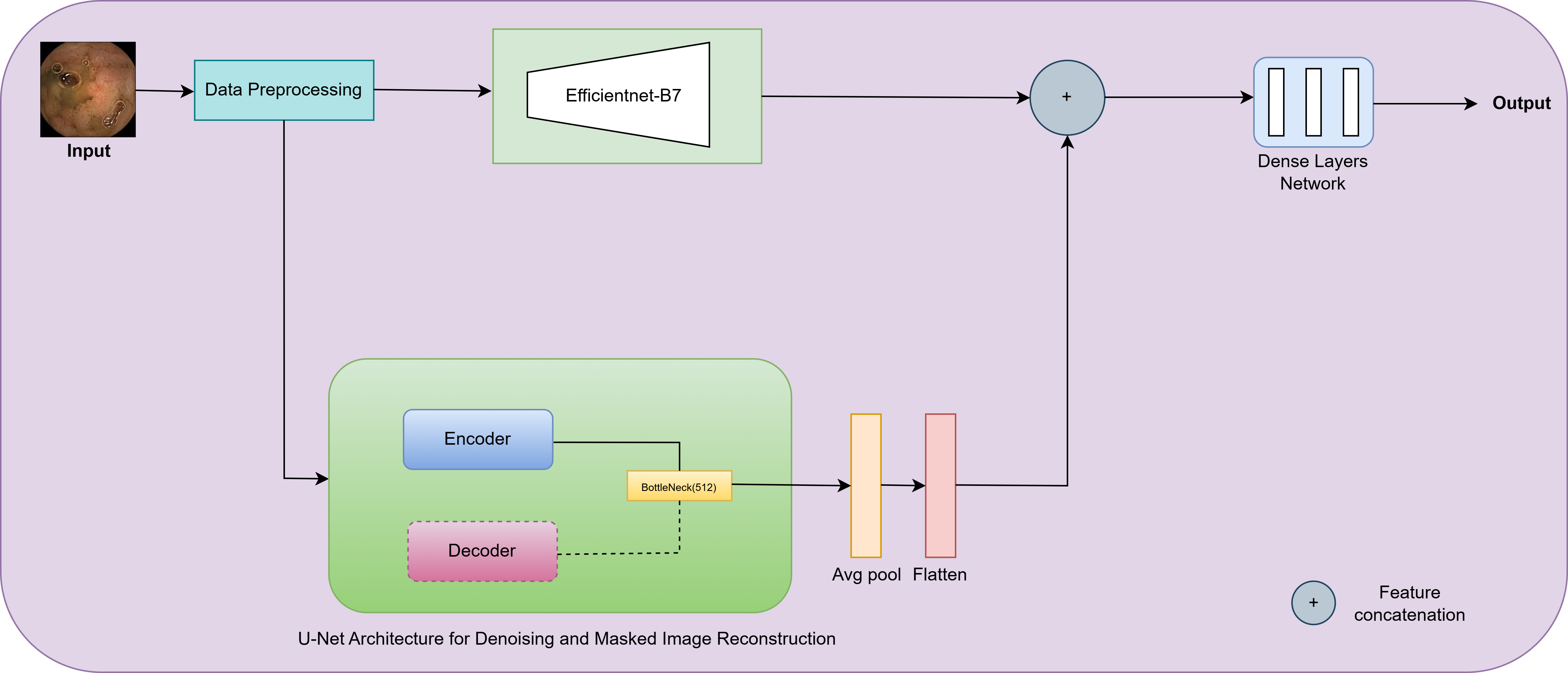}
    \caption{Proposed Block diagram combining EfficientNet(B7) and U-Net Self Supervisied}
    \label{fig:enter-label}
\end{figure}

\begin{figure}[htbp]
    \centering
    \includegraphics[width=0.7\linewidth]{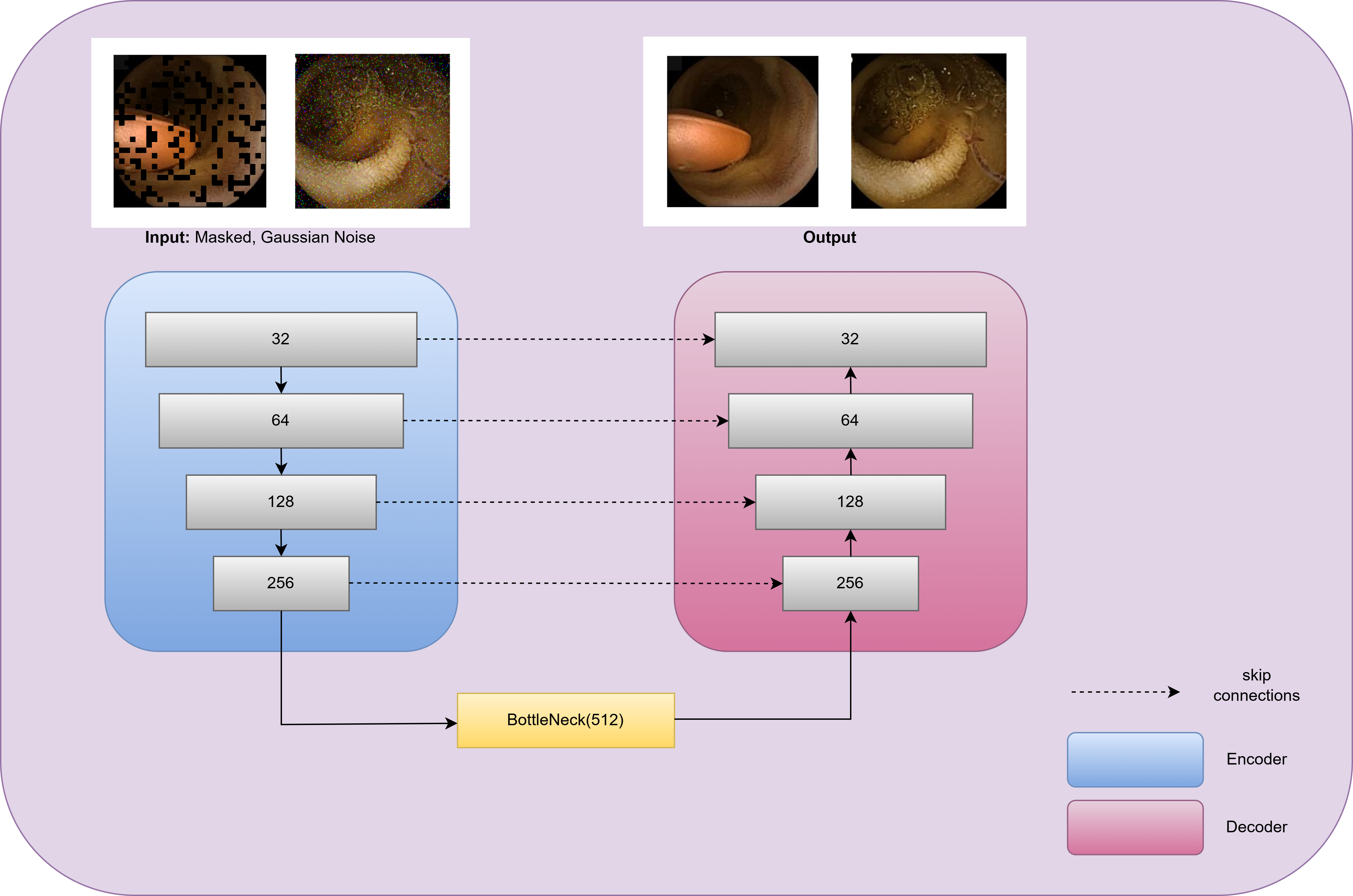}
    \caption{U-Net Architecture for Denoising and Masked Image Reconstruction}
    \label{fig:unet_arch}
\end{figure}

The proposed framework is designed to leverage both self-supervised and supervised learning paradigms to achieve robust and accurate classification of gastrointestinal abnormalities in Video Capsule Endoscopy images. The overall pipeline consists of two major components: (1) a self-supervised U-Net-based masked autoencoder for learning rich visual representations through image reconstruction shown in Figure~\ref{fig:unet_arch}, and (2) a supervised EfficientNet-B7 backbone for discriminative feature extraction. These two streams are integrated through feature fusion and followed by a fully connected classifier. The comprehensive workflow of our proposed dual-branch network is illustrated in Figure~\ref{fig:enter-label}.

\subsection{Self-Supervised Reconstruction Module (U-Net)}

The first stage of the framework focuses on learning meaningful representations from unlabeled VCE images. Inspired by masked autoencoders and denoising autoencoders, we design a U-Net architecture that is trained to reconstruct clean images from corrupted inputs. The motivation behind this approach is that by forcing the network to recover missing or noisy regions, it must learn high-level semantic and structural information about the gastrointestinal anatomy and potential abnormalities.

We explore two corruption strategies: random patch masking and Gaussian noise injection. In the masking approach, random patches of the input image are removed or replaced with zeros, and the network is trained to predict the missing pixels. In the Gaussian noise approach, noise sampled from a Gaussian distribution is added to the input image, and the network is trained to denoise and reconstruct the original image. Both strategies encourage the model to learn robust features; however, empirical evaluation showed that Gaussian noise removal produced superior reconstruction quality and more stable training, likely due to its ability to preserve global context while still challenging the network.

The U-Net architecture consists of an encoder with four convolutional blocks containing 32, 64, 128, and 256 channels, respectively, followed by a bottleneck layer with 512 channels. Each block includes convolution, batch normalization, and ReLU activation, along with max pooling for downsampling. The decoder mirrors the encoder structure with upsampling layers and skip connections that concatenate corresponding encoder features to preserve spatial information. The output layer uses a sigmoid activation to generate the reconstructed image.

The model is trained using a reconstruction loss, such as mean squared error, between the output and the clean target image. Through this process, the encoder learns to capture high-level representations that are invariant to noise and artifacts commonly present in VCE images.

\begin{figure}[htbp]
    \centering
    \includegraphics[width=\linewidth]{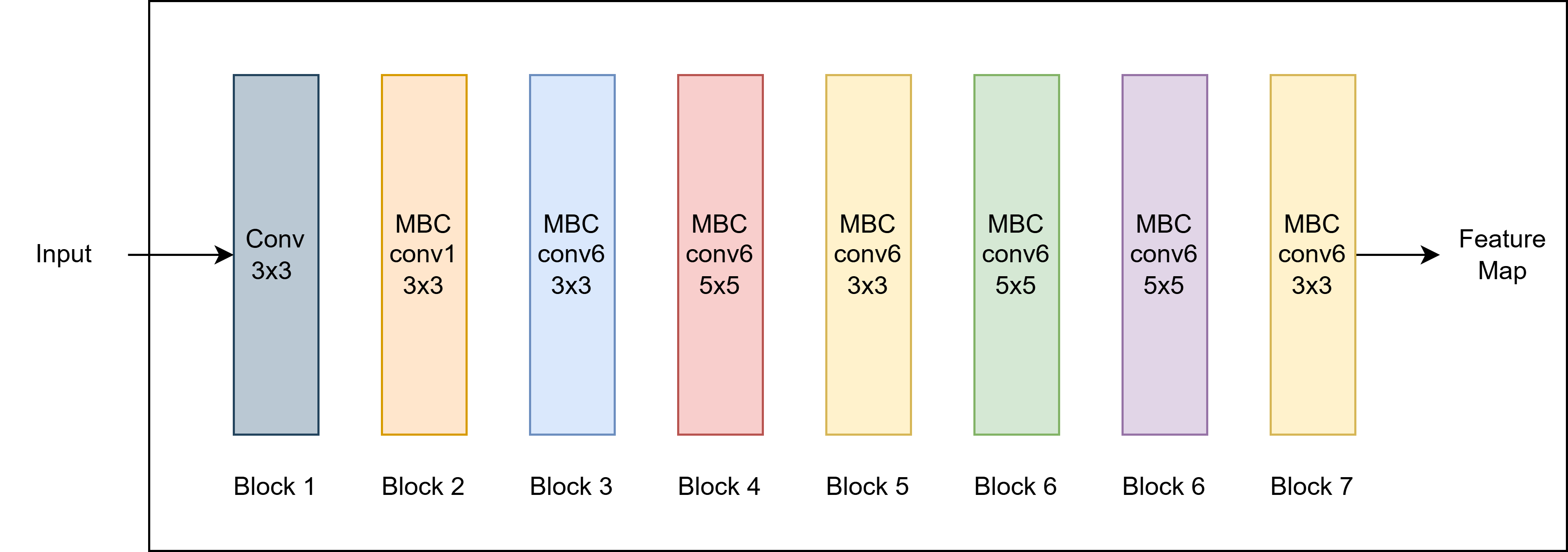}
    \caption{Efficientnetb7 model architecture}
    \label{fig:efficient_block}
\end{figure}

\subsection{Supervised Feature Extraction with EfficientNet-B7}

In parallel to the self-supervised branch, we employ EfficientNet-B7 as a supervised feature extractor. EfficientNet-B7 is chosen due to its strong performance and ability to capture both fine-grained details and global context. The network is initialized with ImageNet-pretrained weights and fine-tuned on the VCE dataset. The global average pooling layer output is used as the EfficientNet feature representation.

\subsection{Feature Aggregation and Classification}

After training the U-Net autoencoder, the encoder part is used to extract features from input images. These features are passed through average pooling and flattened to form a feature vector. This vector is then concatenated with the EfficientNet-B7 feature vector, resulting in a fused representation that combines reconstruction-based and discriminative features.

The fused feature vector is fed into a series of fully connected layers with ReLU activation and dropout for regularization. The final layer uses a softmax activation to output class probabilities for the ten target classes. This feature-level fusion allows the model to leverage complementary information from both architectures, improving robustness and generalization.
All models are trained using the Adam optimizer with a learning rate of 0.0001 and a batch size of 256. Categorical cross-entropy is used as the loss function. Data augmentation techniques such as rotation, flipping, and brightness adjustment are applied to reduce overfitting. The self-supervised U-Net is trained first, followed by supervised fine-tuning of the fusion model.

\section{Experimental Results and Analysis}

In this section, we present a comprehensive evaluation of the proposed self-supervised U-Net and EfficientNet-B7 fusion framework on the Capsule Vision 2024 Challenge dataset. Our experiments are designed to assess the effectiveness of self-supervised reconstruction learning, feature fusion, and the overall impact of the proposed architecture on multi-class abnormality classification in VCE images.

\subsection{Evaluation Metrics}

Given the highly imbalanced nature of the dataset, relying solely on overall accuracy can be misleading. Therefore, we report precision, recall, F1-score, macro average, weighted average, and overall accuracy. Macro-averaged metrics treat all classes equally and provide insight into performance on minority classes, while weighted averages account for class frequencies. These metrics together provide a balanced view of model performance.

We evaluated several models and determined the best one based on balanced accuracy. This approach allowed us to select the most effective model for our task. Among the models evaluated, the Efficient Fusion U-Net with Attention achieved the highest accuracy of 0.929,  demonstrating its ability to effectively interpret the complexities of the data. The Efficient model followed closely with an accuracy of 0.919, while the standard U-Net model showed a lower accuracy of 0.814. Additionally, the Efficient Fusion U-Net model achieved an accuracy of 0.940, further confirming its superior performance. Table~\ref{tab:model_comparison} shows the results of the Efficient with U-Net fusion.

\subsection{Quantitative Results and Comparative Analysis}

We evaluate the following models:
\begin{itemize}

\item  \textbf{U-Net (standalone classifier)}: U-Net encoder followed by dense layers for classification.

\item  \textbf{EfficientNet-B7}: Fine-tuned EfficientNet-B7 using only supervised learning.

\item  \textbf{Efficient Fusion U-Net with Attention}: Feature fusion with attention mechanism.

\item  \textbf{Efficient Fusion U-Net (Proposed)}: Feature-level concatenation of U-Net encoder and EfficientNet-B7 features without attention.
\begin{table}[ht]
    \centering
    \caption{Benchmarking against baseline models. Our proposed method demonstrates superior accuracy.}
    \label{tab:model_comparison}
    \begin{tabular}{lc}
        \toprule
        \textbf{Model Architecture} & \textbf{Accuracy} \\
        \midrule
        U-Net (Standard)                       & 0.814 \\
        EfficientNet (Baseline)                & 0.919 \\
        Efficient Fusion U-Net with Attention  & 0.929 \\
        \textbf{Efficient Fusion U-Net (Ours)} & \textbf{0.940} \\
        \bottomrule
    \end{tabular}
\end{table}

\end{itemize}
This ablation-style comparison allows us to isolate the contribution of self-supervised learning and feature fusion.

The classification results demonstrate a clear advantage of the proposed fusion approach. The standalone U-Net model achieves an accuracy of 0.814, which, while reasonable, indicates that reconstruction-based features alone are insufficient for high-level discrimination. EfficientNet-B7 improves performance significantly, achieving an accuracy of 0.919, highlighting the strength of supervised discriminative learning.

The Efficient Fusion U-Net with attention achieves an accuracy of 0.929, suggesting that combining features from both networks is beneficial. However, the highest performance is achieved by the proposed Efficient Fusion U-Net model, which reaches an accuracy of 0.940. This confirms that simple feature concatenation is sufficient to exploit complementary information from both branches and that attention mechanisms are not strictly necessary in this context.

\subsection{Class-Wise Performance Analysis}

A detailed per-class analysis reveals important insights. The model achieves very high precision and recall for critical classes such as ulcer (F1 = 0.982), normal (F1 = 0.982), and worms (F1 = 0.971). These results are particularly encouraging given the clinical importance of correctly identifying ulcers and parasitic infections.
\begin{table}[ht]
    \centering
    \caption{Detailed classification metrics for Efficient Fusion U-Net. The separation of class-wise performance highlights the model's robustness even on minority classes like `Worms'.}
    \label{tab:class_metrics}
    \setlength{\tabcolsep}{6pt} 
    \begin{tabular}{l ccc r} 
        \toprule
        \textbf{Class} & \textbf{Precision} & \textbf{Recall} & \textbf{F1-Score} & \textbf{Support} \\
        \midrule
        Angioectasia     & 0.865 & 0.813 & 0.838 & 497 \\
        Bleeding         & 0.840 & 0.822 & 0.831 & 359 \\
        Erosion          & 0.785 & 0.764 & 0.774 & 1,155 \\
        Erythema         & 0.614 & 0.535 & 0.572 & 297 \\
        Foreign Body     & 0.911 & 0.844 & 0.876 & 340 \\
        Lymphangiectasia & 0.888 & 0.854 & 0.871 & 343 \\
        Normal           & 0.978 & 0.986 & 0.982 & 12,287 \\
        Polyp            & 0.692 & 0.752 & 0.721 & 500 \\
        Ulcer            & 0.989 & 0.976 & 0.982 & 286 \\
        Worms            & 0.944 & 1.000 & 0.971 & 68 \\
        \midrule 
        \multicolumn{3}{l}{\textit{Overall Accuracy}} & \textbf{0.940} & 16,132 \\
        \textit{Macro Avg} & 0.851 & 0.835 & 0.842 & 16,132 \\
        \textit{Weighted Avg} & 0.939 & 0.940 & 0.939 & 16,132 \\
        \bottomrule
    \end{tabular}
\end{table}

Classes such as angioectasia, bleeding, and lymphangiectasia also show strong performance, with F1-scores above 0.83. This indicates that the model is effective at capturing subtle vascular and mucosal patterns.

The performance on erythema and polyp is comparatively lower, with F1-scores of 0.572 and 0.721, respectively. This is likely due to the subtle visual appearance of erythema and the relatively small size and shape variability of polyps, combined with limited sample sizes. Nonetheless, the recall values suggest that the model is able to detect a significant portion of these cases, which is clinically valuable.


The macro-averaged F1-score of 0.842 demonstrates that the model maintains good balance across classes, while the weighted F1-score of 0.939 reflects strong overall performance dominated by the majority class. Importantly, the improvement in macro metrics compared to baseline models indicates that self-supervised feature learning and fusion help mitigate class imbalance to some extent.

\section{Discussion}
Our experiments indicate that combining reconstruction-based representations with discriminative CNN features offers a substantial advantage over standalone models. This is particularly valuable in medical imaging, where obtaining large-scale annotated datasets remains a constraint.

A key finding from our study is the superiority of Gaussian noise-based reconstruction over standard masked patch modeling. Denoising has been shown to be more stable and effective for downstream prediction. This is likely due to the inherent characteristics of VCE images, which often exhibit motion blur, air or fluid bubbles, and inconsistent lighting. By directing the U-Net model to improve the image, the encoder learns to recognize fundamental anatomical structures rather than just understanding surface-level textures.

Regarding feature fusion, we observed that a simple concatenation strategy outperformed complex attention-based mechanisms. EfficientNet-B7 provides a strong global semantic context, while self-supervised U-Net contributes structural detail. The fact that attention did not produce better results suggests that the features from both branches are already highly discriminative; adding attention layers likely introduced unnecessary complexity rather than refining the feature space.

The proposed model shows improved macro-averaged performance; however, minority classes such as erythema and polyps remain difficult to classify perfectly due to their subtle visual characteristics and the limited number of training samples. Nevertheless, the increased recall for these classes shows that self-supervised pretraining helps the model generalize better to rare cases than supervised learning alone.

From a clinical point of view, the high sensitivity to critical medical conditions such as bleeding and ulcers shown significant improvement. The system is well-suited to act as a triage tool flagging high-risk frames for priority review rather than a standalone diagnostician.

Limitations of the current framework include its lack of temporal modeling. VCE abnormalities often span multiple consecutive frames, and our model currently treats each frame in isolation. Additionally, the computational weight of EfficientNet-B7 could be a hurdle for real-time edge deployment. Future iterations could explore lighter backbones or knowledge distillation to address this.

\section{Conclusion}
This paper presented a dual-branch framework that effectively integrates self-supervised learning with supervised feature fusion for VCE abnormality classification. By leveraging a U-Net trained on a denoising pretext task, we were able to extract robust anatomical features from unlabeled data, which complemented the semantic power of EfficientNet-B7.

With a validation accuracy of 94\% on the Capsule Vision 2024 dataset, our Efficient Fusion U-Net outperforms both standalone baselines and attention-based variants. The results confirm that self-supervision is a potent strategy for overcoming data scarcity in medical domains. Furthermore, the success of simple feature fusion highlights that well-represented features are often more important than complex aggregation modules.

Looking ahead, we aim to extend this work by incorporating temporal information to exploit frame-to-frame consistency. We also plan to investigate class-balancing loss functions to further improve performance on rare pathologies. Ultimately, this approach represents a step toward more reliable, automated decision-support systems for gastrointestinal diagnostics.

\section{Ablation Study: Effectiveness of Self-Supervision}

One of the key findings is that self-supervised pretraining of the U-Net significantly improves downstream classification. When the U-Net is trained only in a supervised manner, performance is noticeably lower. The denoising and reconstruction task forces the encoder to learn robust, anatomy-aware representations, which complement the discriminative features learned by EfficientNet-B7. This synergy is a major contributor to the performance gains observed. In general, the experimental results validate the effectiveness of the proposed framework and demonstrate its potential for clinical implementation in the real-world.

\section*{Acknowledgments}
As participants in the Capsule Vision 2024 Challenge, we fully comply with the competition's rules as outlined in \cite{capsule2024}. Our AI model development is based exclusively on the datasets provided in the official release in \cite{dataset2024}.

%
%
%

\begin{thebibliography}{8}

\bibitem{okeke2025}
Okeke, S., et al.: Precision enhancement in wireless capsule endoscopy: A Review. Front. Artif. Intell. \textbf{8}, 1529814 (2025). \doi{10.3389/frai.2025.1529814}

\bibitem{soffer2025}
Soffer, S., et al.: Recent technological advances in video capsule endoscopy: a comprehensive review. Clin. Endosc. \textbf{58}(1), 1--12 (2025). \doi{10.5946/ce.2025.135}

\bibitem{togacar2022}
Togacar, M., Cömert, Z., Ergen, B.: A Novel Multi-Feature Fusion Method for Classification of Gastrointestinal Diseases Using Endoscopy Images. Diagnostics \textbf{12}(10), 2316 (2022). \doi{10.3390/diagnostics12102316}

\bibitem{kundu2022}
Kundu, R., et al.: Classification of wireless capsule endoscopy images for bleeding using deep features fusion. Sci. Rep. \textbf{12}, 1458 (2022). \doi{10.1109/ICECCME55909.2022.9987916}

\bibitem{khan2022}
Khan, M.A., et al.: Deep Feature Fusion and Optimization-Based Approach for Stomach Disease Classification. Sensors \textbf{22}(7), 2801 (2022). \doi{10.3390/s22072801}

\bibitem{caron2021}
Caron, M., Touvron, H., Misra, I., Jégou, H., Mairal, J., Bojanowski, P., Joulin, A.: Emerging properties in self-supervised vision transformers. In: Proceedings of the IEEE/CVF International Conference on Computer Vision (ICCV). pp. 9650--9660 (2021). \doi{10.1109/ICCV48922.2021.00951}

\bibitem{he2022}
He, K., Chen, X., Xie, S., Li, Y., Dollár, P., Girshick, R.: Masked autoencoders are scalable vision learners. In: Proceedings of the IEEE/CVF Conference on Computer Vision and Pattern Recognition (CVPR). pp. 16000--16009 (2022). \doi{10.1109/CVPR52688.2022.01553}

\bibitem{endovit2024}
Batic, D., et al.: EndoViT: pretraining vision transformers on a large collection of endoscopic images. Int. J. Comput. Assist. Radiol. Surg. \textbf{19}, 13–21 (2024). \doi{10.1007/s11548-024-03091-5}

\bibitem{capsule2024}
Handa, P., et al.: Capsule vision 2024 challenge: Multi-class abnormality classification for video capsule endoscopy. arXiv preprint arXiv:2408.04940 (2024). \doi{10.48550/arXiv.2408.04940}

\bibitem{dataset2024}
Handa, P., et al.: Training and Validation Dataset of Capsule Vision 2024 Challenge. Figshare (2024). \doi{10.6084/m9.figshare.26403469.v1}



\end{thebibliography}
%

\end{document}